# DEANet: Decomposition Enhancement and Adjustment Network for Low-Light Image Enhancement

Yonglong Jiang[1], Liangliang Li[2], Yuan Xue[3], and Hongbing Ma[1]*

*Abstract*—Images obtained under low-light conditions will seriously affect the quality of the images. Solving the problem of poor low-light image quality can effectively improve the visual quality of images and better improve the usability of computer vision. In addition, it has very important applications in many fields. This paper proposes a DEANet based on Retinex for low-light image enhancement. It combines the frequency information and content information of the image into three sub-networks: decomposition network, enhancement network and adjustment network. These three sub-networks are respectively used for decomposition, denoising, contrast enhancement and detail preservation, adjustment, and image generation. Our model has good robust results for all low-light images. The model is trained on the public data set LOL, and the experimental results show that our method is better than the existing state-of-the-art methods in terms of vision and quality.

## I. Introduction

IN daily life, we may encounter scenes that make us refreshed and pleasant at any time, either during the day with high light or at night with low light. In this case, due to the limitations of the shooting equipment, how to take a high-quality clear image has become a challenge. Professional photographers can obtain clear images by setting high ISO, long exposure and flash, but they also have different shortcomings. High ISO will result in increased noise in the dark and pure colors, which will result in too low Peak Signal to Noise Ratio (PSNR). Long exposure can reduce noise by increasing the amount of light, but there will be thermal noise interference, resulting in poor image quality. The flash will enhance the local brightness of the image, but it is visually unacceptable. So people have proposed to solve this problem through low-light image enhancement. Low-light image enhancement has made great progress in the past few years, but enhancement of high-definition low-light image is still challenging.

The first line of Fig. 1. provides three natural images taken under severe lighting conditions. The first image was taken at sunset (low-light environment) and the result of the imaging of objects under extremely low light. The second image was taken when the sun is rising and the object is backlit when the ambient light is sufficient. The third image was taken under the condition of low daily indoor light. The second line is the enhancement result corresponding to our model.

From the above three enhanced images, it can be seen that the deep learning method has shown superior performance in the task of low-illumination image enhancement. The method can also make hidden details clearly visible and improve the subjective feeling and usability of the current computer vision system at the same time. However, low-light image enhancement networks still face many challenges in practical applications, as follows:

(a) How to use the Retinex theory to separate the appropriate illumination component and reflection component through the deep learning network can effectively improve the contrast of the image.
(b) How to increase the contrast while retaining more details of the image and suppress the expansion of noise at the same time.
(c) How to restore a clear image from the details learned by the neural network.

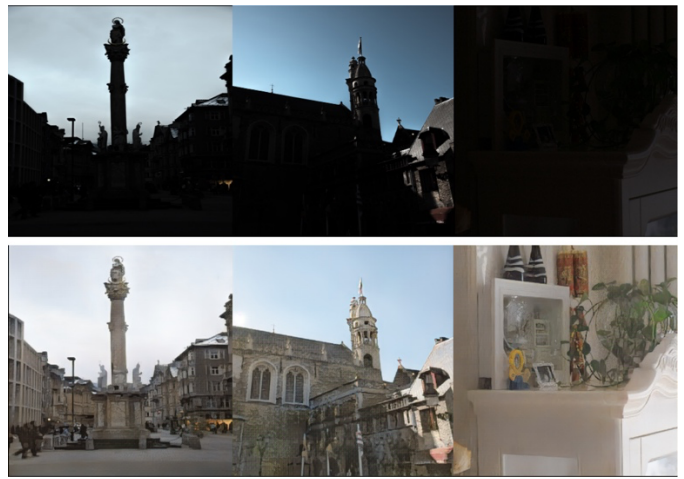

Fig. 1. Top column: three natural images captured under different light conditions. Bottom column: our enhanced results

[1]College of Information Science and Engineering, Xinjiang University, Urumqi 830046, China; [2]Department of Electronic Engineering, Tsinghua University, Beijing 100084, China [3]College of Information Science and Engineering, Xinjiang University, Urumqi 830046, China; Corresponding author: Hongbing Ma Email: hbma@tsinghua.edu.cn

This paper comprehensively considers the above problems and proposes a DEANet deep neural network to solve the above problems.

## A. Previous Works

Many low-light image enhancement schemes have been proposed. The main traditional and contemporary methods will be briefly introduced in the following.

**Traditional methods.** For an image with very dark overall light, when the distribution of each gray value is finely balanced, the amount of information contained in the image becomes very large. On the contrary, when there is only one gray value, the amount of information is quite small. The technical route represented by Histogram Equalization (HE) [1-3] and its following [4-5] maps the grayscale value range of the image to [0,1], which redistributes the image pixel values, so that the numbers of pixels in different grayscale ranges are roughly equal. In this way, the output histogram is balanced, which avoids the over concentration of the grayscale and then improves the contrast of image. Another way is to perform non-linear gamma correction (GC) on each pixel, which can effectively improve the brightness, but ignores the relationship between pixels. The disadvantage of the traditional method is that it only simply enhances the brightness and ignores the real lighting factors, resulting in a huge difference between the visual and the real scene.

**Lighting-based methods.** Different from the traditional method, the lighting-based method is affected by the Retinex theory [6], separating the image into two components, the illumination component and the reflection component. Early attempts using this method are listed below: Single-scale Retinex (SSR) [7], which restricts the smoothing of the illumination component through a single-scale Gaussian filter. Multi-scale Retinex (MSR), which enhances SSR through multi-scale Gaussian filter and color restoration. Based on the MSR, the MSR method with color restoration MSRCR [8] adds a color restoration factor C to adjust the defect of color distortion caused by the contrast enhancement of the local area of the image. BIMEF [9] uses double exposure algorithm for image enhancement. LIME [10] estimates the lighting through pre-suppositions, obtains the estimated lighting through a weighted model, and then uses BM3D [11] as a post-processing. Wang et al. proposed a method called NPE [12], which can enhance the contrast while ensuring the naturalness of the image lighting. Fu et al. [13] proposed a method that enhances low-brightness images by fusing multiple images, but this method will sacrifice the realism of some detail areas of the image. Guo et al. [14] selected the maximum value in each pixel channel to initialize the image illumination map, and then refine the initial illumination map by adding a priori structure, and finally synthesize the enhanced image according to the Retinex theory [6]. In SRIE [15], the author obtained better reflective layer and illumination layer by using a weighted variational model.

**Deep learning methods.** With the popularization and application of deep learning, many low-level vision tasks can be improved by using deep learning models. For example, Xie et al. [16] and Zhang et al. [17] are used to remove noise interference, Dong et al. [18] is used for super-resolution, Deng et al. [19] is used to remove artifacts, and Cai et al. [20] is used to remove impurities. For the task of low-light image enhancement, LLNet [21] built a deep learning network that uses stacked sparse denoising autoencoders to enhance and denoise the low light images at the same time. MSR-net [22], imitating the process of traditional MSR to directly learn the end-to-end mapping from low-light images to normal light images. Wei et al. [23] designed a deep learning network RetinexNet that combines Retinex theory with DeepCNN to estimate and adjust the illumination map. This method achieves image contrast enhancement, and then uses BM3D for post-processing to achieve denoising. MBLLEN [24] extracts the rich features of the image to different levels through the CNN convolutional layer, uses multiple subnets for simultaneous enhancement, and finally fuses the multi-branch output results into the final enhanced image. Lim et al. [25] proposed a deep stacked Laplacian restorer (DSLR) to restore global illumination and local details from the original input. GLAD [26] first calculates the global illumination estimate of the low-light input, adjusts the illumination under the guidance of the illumination estimate, and then supplements the details by cascading with the original input.

In addition, there are some methods different from the above. Ying et al. [27] used the camera response model for weak illumination image enhancement. Lore et al. [28] proposed a stacked sparse denoising autoencoder for image contrast enhancement and denoising. RRDNet [29] proposed a new three-branch convolutional neural network, which decomposes the input image into three components of illumination, reflection and noise. It does not require paired data sets as data drivers, and iteratively minimizes a special designed loss function to update. Guo et al. [30] proposed a lightweight network Zero-DCE, which transformed the image enhancement problem into a curve estimation problem. Jiang et al. [31] proposed a GAN-based network for low-light image enhancement. It used unpaired images for training for the first time. However, the existing models still have problems, such a the regional degradation and so on. For example, many models have over exposure, amplified noise and color distortion, resulting in a poor image quality. Besides, none of the above methods comprehensively consider the influence of noise on the image in terms of image frequency and content. The model proposed in this paper will solve the above problems.

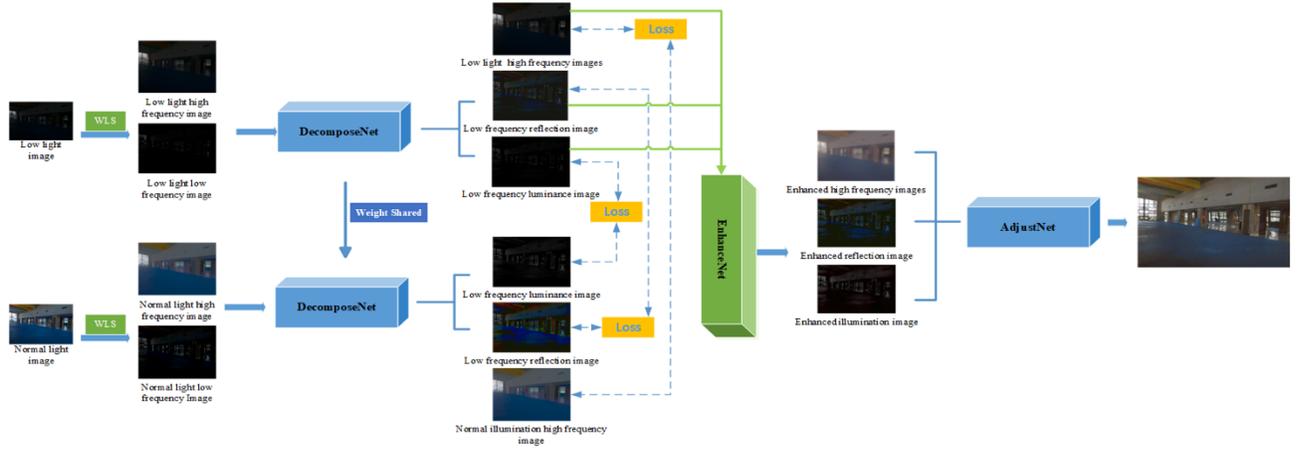

Fig. 2. The proposed DEANet architecture.

### B. Our Contributions

A new deep learning network is proposed in this paper, which solves the problem of low-light image enhancement. The contribution of this paper can be summarized as follows:
- Affected by the Retinex theory, a DEANet is proposed. It uses the high frequency of the image to denoise, and the low frequency of the image is combined with the content of the image to adjust the brightness of the details.
- In the enhanced network, a new network structure is proposed. It combines DenseNet and Unet network, which can clearly restore the details of the image.
- In the final adjustment network, a network structure combining ResNet and Unet network is proposed, which can clearly restore the color of the image.
- In the decomposition network, enhancement network and adjustment network, different loss functions are designed to ensure the restoration of image features.
- We try to combine the adjustment network with the enhancement network and make adjustment network use its backward to guide the learning of the enhancement network. This method can effectively accelerate the convergence speed of the enhanced network.
- We have conducted a lot of experiments to ensure the effectiveness of our design and its superiority comparing to other existing technical solutions.

## II. METHODOLOGY

An excellent low-light image enhancement model should be able to effectively restore the details of the image while trying to avoid the degradation problem hidden in the dark. In order to achieve this goal, a new deep network structure is established. As shown in Fig. 2, the network has three branches, which are used to suppress noise, reduce the degradation of object reflection and adjust the light. The detailed information about the network will be introduced in the following.

### A. Thinking and Hypothesis

*1) Thoughts on Retinex Theory*

Inspired by the Retinex theory, the simple decomposition method of $I = R \circ L$ ignores the interference of noise, and the real image cannot ignore the interference of noise. So the decomposition idea of $I = R \circ L + N$ is proposed, R is the reflection component of the object, L is the brightness component of the object, and N is the noise. ∘ means to multiply according to the element. This decomposition method can decouple the image map to the image space into three smaller subspaces, which can make regularization learning easier.

*2) Data-driven Decomposition*

The decomposition of the image is generally carried out through carefully designed constraints, and the reflection images and brightness images are obtained after decomposition, but such carefully designed constraints cannot be adapted to various occasions. Therefore, data-driven decomposition of images is used to solve this problem. In the training phase, the decomposition network uses pairs of low-light images and normal-light images at the same time, and afterwards obtains the paired light map and reflection map after decomposition. Because in actual situations, there is a phenomenon that low-light image degradation is usually more serious than normal-light image degradation. This phenomenon is also transferred to the reflected image and the brightness image, so the enhancement module is adopted to learn the mapping relationship which ensures that the degradation between the reflected image and the illuminated image is eliminated.

*3) Image Adjustment and Restoration*

The $I = R \circ L + N$ thoughts guide the restoration of images. An auxiliary network is proposed to adjust and restore the images. In the real life, even in images with normal illuminance, noise still exists, so we get a low-noise image, the reflection images and illuminance images obtained by the enhancement network to guide the restoration of the image.

## B. DEANet

Based on the above thoughts and assumptions, a deep neural network called DEANet is designed to ensure the details of the image while restoring the brightness of the low-light image. The specific structure is shown in Fig. 2.

This network consists of three subnets: a Decom-Net, a Enhance-Net, and a Adjust-Net, which performs decomposing, contrast enhancing, and detail Adjusting respectively. The Decom-Net decomposes the low-light image into an illumination map and a reflectance map based on the Retinex theory. The Enhance-Net aims to suppress the noise in the high frequency map, learn about light mapping and resolve reflection degradation. Subsequently, the illumination map, the reflection map and high frequency map obtained by Enhance-Net will be sent to the Adjust-Net to improve image contrast and reconstruct details.

In the following section, these three modules will be described in detail from a functional perspective.

### 1) Decompose the network

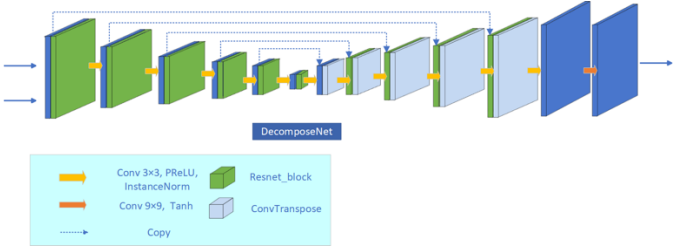

Fig. 3. DecomposeNet

First of all, most of the noise in the image is in the high-frequency information of the image, so a weighted least squares (WLS) is used. This edge-preserving filter can separate low light pictures and normal light pictures into high frequency pictures and low frequency pictures as close as possible to the original picture. The high frequency image stores the noise of the image and the boundary information of the image. Low frequency images store the details and the brightness that needs to be adjusted of the image. Without the guidance of real lighting and other information, the pre-design the constraint function for separation can be hardly obtained. Fortunately, using a data-driven decomposition network [23], the paired images [$I_{low}$, $I_{high}$] with different exposure configurations can be separated to get the lighting [$L_{low}$, $L_{high}$] in various situations in the image. According to the Retinex theory, when degradation is considered, the reflectivity of the object remains unchanged in different scene images. The decomposed reflectivity [$R_{low}$, $R_{high}$] is adjusted as close as possible. While the degradation problem is ignored, the reflectivity should be as same as possible. Several loss functions are constructed to ensure the correctness of the separation. Firstly, the $L_r = \| R_{low} - R_{high} \|_1$ loss is used to ensure that the reflectivity of the object is as same as possible. $R_{low}$ represents the reflectivity of low-light images, $R_{high}$ represents the reflectivity of the image under normal lighting, $\|\ \|$ means L1 loss. Secondly, a pair of low-light images and normal-light images is used to get the maximum values of the pixels on their RGB three channels as the respective brightness map. The single-channel brightness map and RGB image are spliced together in the channel dimension and input to the decomposition network. Using $L_{recon\_low} = \| R_{low} \circ L_{low} - I_{low} \|_1$ and $L_{recon\_high} = \| R_{high} \circ L_{high} - I_{high} \|_1$ to ensure the correct decomposition of low-light images and normal-light images, $L_{low}$ represents the light component of a low-light image, and $L_{high}$ represents the light component of a normal-light image. Finally, $L_{recon\_low\_mutal} = \| R_{high} \circ L_{low} - I_{low} \|_1$ and $L_{recon\_high\_mutal} = \| R_{low} \circ L_{high} - I_{high} \|_1$ are used to ensure that the reflectance R is as smooth as possible. The loss function of the decomposition layer can be expressed as:

$$L_{decom} = 0.01 \times L_r + L_{recon\_low} + L_{recon\_high} \\ + 0.001 \times (L_{recon\_low\_mutal} + L_{recon\_high\_mutal}) \quad (1)$$

The decomposition layer network consists of two parts. The first part uses a WLS filter, and the second part is a typical combined 6-layer UNET and ResNet network. This combined network adopts a long and short hop hybrid connection method, which can transmit information to the greatest extent and reduce information loss, thereby ensuring the details of the image during up sampling.

### 2) Enhance the Network

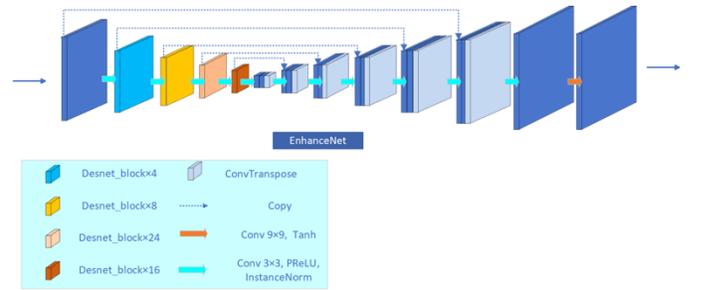

Fig. 4. EnhanceNet

On the enhanced network, three DenseNet networks with excellent performance are designed. The high-frequency image pair [$HF_l$, $HF_h$] obtained by the first network through WLS. Under the constraint of using L1 loss function $L_{HF} = \| HF_l - HF_h \|_1$, the enhancement network can effectively learn the mapping relationship from the noise of low-light images to the noise of normal light images, which not only eliminates a lot of noise but also learns the contour features of the objects in the image. The second network is the reflection image pair [$LF_{low\_r}$, $LF_{high\_r}$] obtained by decomposing the network. Under the constraint of using L1 loss function $L_{en\_r} = \| LF_{low\_r} - LF_{high\_r} \|_1$, the enhancement network can effectively learn the mapping relationship between low-light illumination components and normal illumination components. This mapping relationship can adjust the brightness of the light, which results in the low light picture is mapped to the normal brightness of the light. Finally, the loss function of the enhancement layer can be expressed as:

$$L_{enhance} = L_{HF} + L_{en\_r} + L_{en\_l} \quad (2)$$

The enhancement layer contains three branch networks and each branch network is a new network formed by the combination of DenseNet and UNet. This new network uses DenseNet's powerful information transmission function, and at the same time, through DenseNet's conversion layer, it transmits information while reducing channels, and transmits the information to the corresponding layer during upsampling. This operation keeps the image to the greatest extent possible and reduces losses of detailed information.

*3) Adjust the Network*

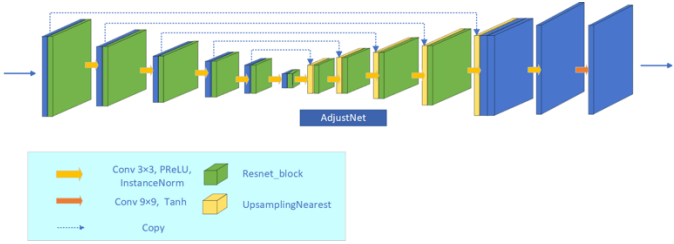

Fig. 5. AdjustNet

To adjust the network, a structure similar to that of the decomposition network is used. Then according to $I = R \circ L + N$, the three images obtained from the enhanced network are combined and sent to the adjust network. In the adjustment network, the content loss function is used. It uses the pre-trained VGG19 network to extract feature pairs [$Feature_{gen}$, $Feature_{real}$] from the generated image and the reference image at the same time. And the feature pairs perform L1 loss $L_{content} = \| Feature_{gen} - Feature_{real} \|_1$, used to ensure the similarity of features. The adjustment network also uses the generated image and the reference image to perform L1 loss $L_{colour} = \| Image_{res} - Image_{nor} \|_1$ on [$Image_{res}$, $Image_{nor}$] to ensure the consistency of the color reproduction of the generated image. Through the combination of the adjustment network and the enhancement network, the adjustment network uses its backward to guide the enhancement network to learn more effectively and accelerate the convergence speed of the enhancement network. Finally, the joint loss function of the adjusted network and the enhanced network is:

$$L_{loss} = 0.1 \times L_{enhance} + L_{colour} + L_{content} \quad (3)$$

The regulation network is a typical combined network of 6-layer UNET and ResNet. This combined network adopts a long and short hop hybrid connection method to efficiently establish a related mapping that ensures the restoration of image details and the restoration of colors.

## III. Experiment

### A. Experimental Details

Our experiment is based on the PyTorch framework. The LOL Dataset is used. The experiment can converge quickly after training 30 epochs on three 1080Ti GPU. The best effect is achieved after 200 epochs. We use the Adam optimizer and set a batch size of one. For more implementation details of the network, please refer to our upcoming code.

### B. Comparison with Existing Methods on Real Data Set

We compare our proposed method with the latest existing methods (MSRCR[8], BIMEF[9], LIME[10], RetinexNet[23], MBLLE[24], DSLR[25], GLAD[26], RRDNet[29], ZeroDCE[30], EnlightenGAN(EG)[31]) on 4 public data sets (LOL, LIME, NPE, MEF). Our model is trained on the LOL Dataset. The LOL Dataset captures 500 pairs of real low/normal light images by changing the camera's exposure time and ISO. This is the only existing real low/normal light image data set used for low-light image enhancement (Sid data set is used for very low light image enhancement). We use the five most commonly used indicators for image quality evaluation: PSNR, SSIM (structural similarity)[32], FSIM (feature similarity)[33], MAE (mean absolute error) and GMSD (gradient magnitude similarity deviation)[34] to comprehensively evaluate our results. Our proposed model has achieved excellent results on those three indicators. But in many cases, the leading evaluation index does not mean that the visual perception is also good. While our visual perception is obviously better than others. The visual effects are shown in Fig 6. The specific evaluation values are shown in TABLE I.

TABLE I
QUANTITATIVE EVALUATION OF LOW-LIGHT IMAGE ENHANCEMENT METHODS ON THE LOL DATASET

| Methods | PSNR | SSIM | FSIM | MAE | GMSD |
|---|---|---|---|---|---|
| MSRCR | 13.964 | 0.514 | 0.827 | **0.046** | 0.151 |
| LIME | 16.758 | 0.564 | 0.850 | 0.097 | 0.122 |
| BIMEF | 13.875 | 0.577 | 0.907 | 0.103 | 0.085 |
| RetinexNet | 16.774 | 0.559 | 0.759 | 0.153 | 0.148 |
| MBLLEN | 18.897 | 0.755 | 0.926 | 0.123 | 0.124 |
| GLAD | 19.718 | 0.703 | 0.923 | 0.129 | 0.112 |
| DSLR | 14.978 | 0.668 | 0.856 | 0.253 | 0.170 |
| RRDNet | 10.998 | 0.456 | 0.823 | 0.341 | 0.194 |
| ZeroDCE | 14.584 | 0.611 | 0.912 | 0.162 | 0.086 |
| EG | 17.239 | 0.678 | 0.911 | 0.087 | **0.084** |
| Ours | **21.260** | **0.798** | **0.938** | 0.078 | 0.103 |

Notes: The best results are highlighted in bold.

It can be seen that the PSNR and SSIM of our method on the LOL Dataset are better than the existing state-of-the-art methods. The proposed DEANet obtains the best performance, with an average PSNR score of 21.260dB and a SSIM score of 0.798. It exceeds the second-best method (MBLLE) by 2.363dB on PSNR and 0.043 on SSIM. It exceeds the third-best method (GLAD) by 1.542dB on PSNR and 0.095 on SSIM. It exceeds the fourth-best method (EG) by 4.021dB on PSNR and 0.120 on SSIM. It exceeds the fifth-best method (DSLR) by 6.282dB on PSNR and 0.130 on SSIM. The MSRCR, LIME, RetinexNet, RRD and ZeroDCE are far less effective than ours. The visual comparison is shown in the Fig.3. It can be seen that some methods based on Retinex theory (such as LIME, RetinexNet) can blur details or amplify noise. The enhancement results generated by our method can not only improve the local and global contrast with clearer details, but also effectively remove the noise in the image. These effects are well proved in the experimental results. Please enlarge the image to compare more details.

The LIME, NPE, and MEF are often used as benchmark datasets for the evaluation of low-light image enhancement methods. These datasets only contain low-light images, and the PSNR and SSIM cannot be used for quantitative evaluation. Therefore, we use the reference-free image quality evaluation NIQE to evaluate the performance of our method. The results are shown in TABLE II.

TABLE II
QUANTITATIVE COMPARISON ON LIME, NPE, AND MEF DATASETS IN TERMS OF NIQE

| Metric | NIQE | | |
|---|---|---|---|
| Datasets | LIME-data | NPE-data | MEF-data |
| LIME | 4.154 | 4.262 | 3.715 |
| BIMEF | 3.816 | 4.196 | 3.423 |
| RetinexNet | 4.597 | 4.567 | 4.475 |
| MBLLEN | **3.654** | 4.073 | 4.999 |
| GLAD | 4.128 | 3.969 | **3.334** |
| ZeroDCE | 3.912 | **3.667** | 4.024 |
| EnlightenGAN | 3.719 | 4.113 | 3.575 |
| Ours | 3.890 | 3.771 | 3.574 |

### C. Ablation Experiment

Firstly, we conduct experiments on the three network modules separately based on the LOL Dataset. We remove our adjustment network and enhancement network respectively to prove the effectiveness of the adjustment network and the enhancement network for our model. Removing the conditioning network and enhancing the network will significantly reduce the performance of our model. The result is shown in Fig 7.

Secondly, we conducted experiments on the depth of the three networks. After reducing the depth of the network (the size of the deepest feature map of the network after reduction is 12*12, the size of the deepest feature map of the existing network is 6*6), the measured PSNR = 20.1, SSIM = 0.745. At the same time, we also test the width of the network. When upsampling, we use PixelShuffle to reduce the width of the network by reducing the number of channels. The measured results are PSNR = 18.7 and SSIM = 0.75. The results show that our existing structure experiments have better results.

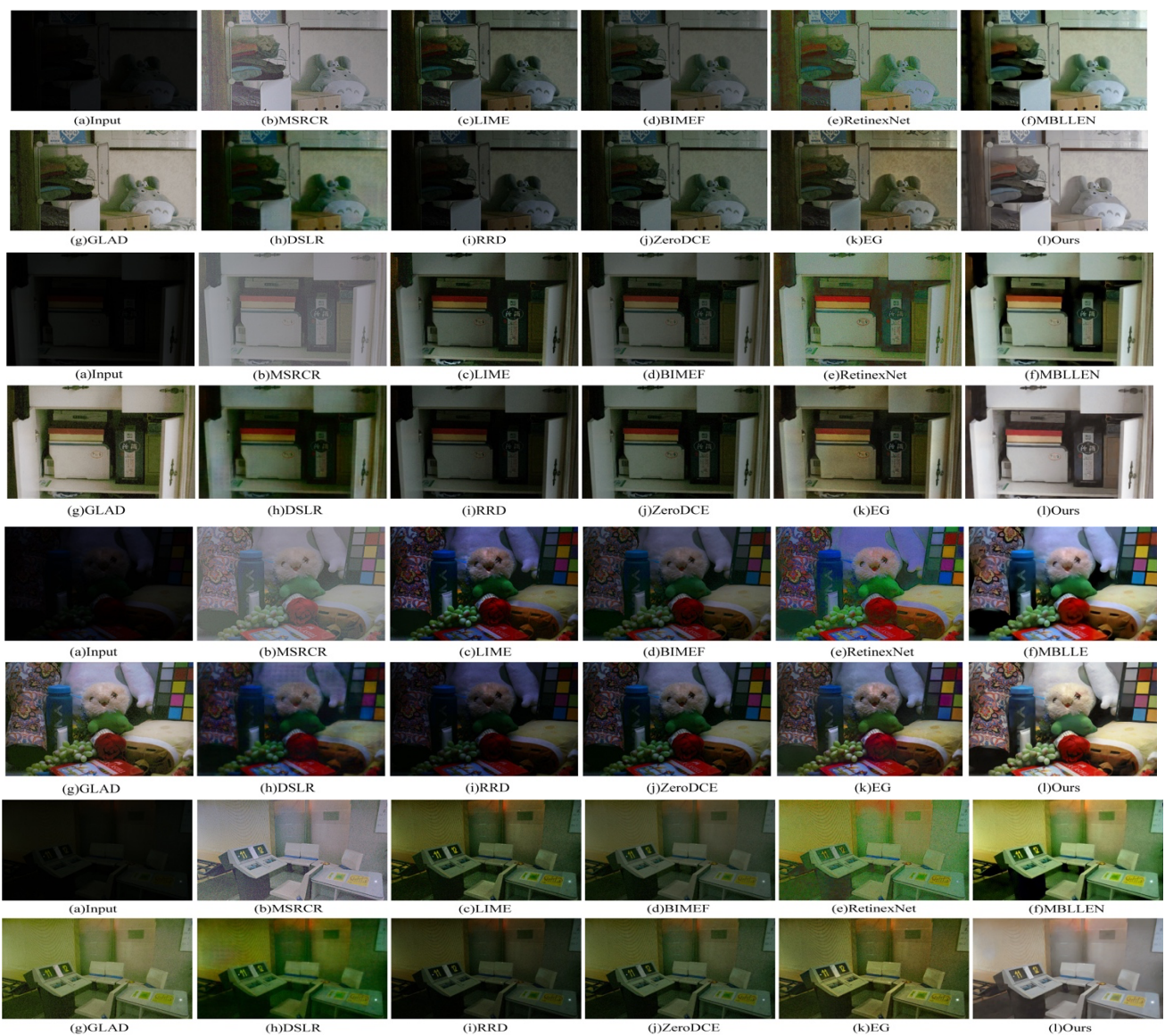

Fig. 6. Visual comparison with state-of-the-art low-light image enhancement methods on the LOL Dataset.

Finally, to adjust the loss function, we replaced the content loss function from VGG19 to DenseNet161 and performed experiments. The results showed that PSNR = 20.1, SSIM = 0.69. Experimental results show that replacing the loss function will result in a decrease in model performance.

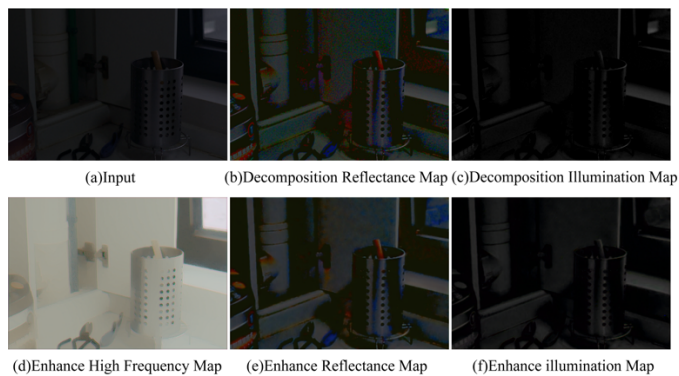

Fig. 7. A(a) Input image (b) (c) Denote images learned by decomposing network. (d) (e) (f) Denote images learned by decompose network and enhance network.

## IV. CONCLUSIONS

This paper is based on Retinex theory and proposes a new end-to-end enhancement network for low-light pictures to normal-light pictures. The network consists of three parts: Decomposition Network, Enhance Network, and Adjust Network. This paper is a preliminary way to guide the recovery of low-light pictures by combining the frequency information and content information of the image used. The enhanced results obtained by our method have better visual effects. It is verified that our scheme of combining image frequency information and content information for low-light image recovery is correct and feasible. The experimental results on the LOL Dataset show that our method can improve the image contrast and good noise suppression and obtain the highest PSNR and SSIM scores, which is far better than other methods. The method in this paper can also be well applied in other fields, such as defogging in pictures, restoring the definition of seabed pictures, and removing artifacts from images.

This paper has achieved excellent results on the LOL Dataset, but at the same time there is room for further improvement on other data sets. We will further explore the recovery of low-light pictures under the guidance of the combination of the frequency and content of the image, so as to further improve our model and make it perform better on other data sets.